\pdfoutput=1

\documentclass[11pt]{article}

\usepackage[preprint]{acl} 
\usepackage{times}
\usepackage{latexsym}
\usepackage[T1]{fontenc}
\usepackage[utf8]{inputenc}
\usepackage{microtype}
\usepackage{inconsolata} 
\usepackage{graphicx}
\usepackage{url}
\usepackage{multirow}

\title{KInIT at SemEval-2024 Task 8: Fine-tuned LLMs for Multilingual Machine-Generated Text Detection}

\author{Michal Spiegel$^{1,2}$ \and Dominik Macko$^1$\\
  $^{1}$ Kempelen Institute of Intelligent Technologies \\
  $^{2}$ Faculty of Informatics, Masaryk University\\
 \texttt{michal.spiegel@intern.kinit.sk}, \texttt{dominik.macko@kinit.sk} \\}

\begin{document}
\maketitle
\begin{abstract}
SemEval-2024 Task 8 is focused on multigenerator, multidomain, and multilingual black-box machine-generated text detection. Such a detection is important for preventing a potential misuse of large language models (LLMs), the newest of which are very capable in generating multilingual human-like texts. We have coped with this task in multiple ways, utilizing language identification and parameter-efficient fine-tuning of smaller LLMs for text classification. We have further used the per-language classification-threshold calibration to uniquely combine fine-tuned models predictions with statistical detection metrics to improve generalization of the system detection performance. Our submitted method achieved competitive results, ranking at the fourth place, just under 1 percentage point behind the winner.
\end{abstract}

\section{Introduction}

Recent large language models (LLMs) are able to generate high-quality texts that are not easily detectable by human readers. A problem arises when such generated texts are misused for academic exams \citep{openai2023gpt4}, plagiarism \citep{Wahle_2022}, disinformation spreading \citep{vykopal2023disinformation}, etc. Therefore, it is crucial to develop automated means to detect machine-generated texts.

SemEval-2024 Task~8 \citep{semeval2024task8} consists of three subtasks: A) binary human-written vs. machine-generated text classification, B) multi-way machine-generated text classification, and C) human-machine mixed text detection. In our work, we have focused on subtask A, especially its multilingual track. It covered 8 known languages for training (Arabic, Bulgarian, Chinese, English, German, Indonesian, Russian, Urdu), multiple domains (e.g., Wikipedia, news, abstracts), and multiple text generators (e.g., GPT-3, ChatGPT, BLOOMZ).

\begin{figure}[!t]
    \centering
    \includegraphics[width=\linewidth, trim=0.2cm 0 2.2cm 0, clip]{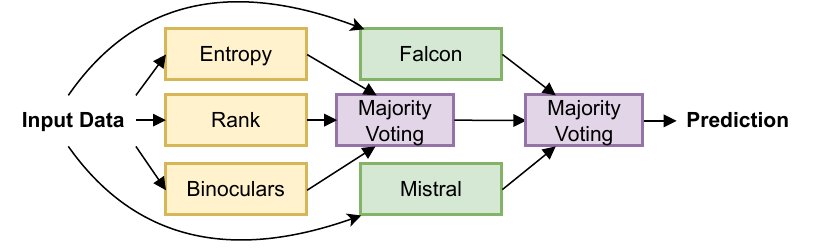}
    \caption{System components overview.}
    \label{fig:architecture}
\end{figure}

During our participation in the shared task, we have explored various alternatives. Our best submitted solution (illustrated in \figurename~\ref{fig:architecture}) combines two fine-tuned LLMs (green-colored) with statistical detection (orange-colored) using a two-step majority voting (purple-colored) based ensemble method. Such a system achieved fourth place in the final leaderboard, with a performance of 95\% in accuracy, within 1 percentage point range behind the winning system. We have published the source code for easier replication purposes\footnote{\url{https://github.com/kinit-sk/semeval-2024-task-8-machine-text-detection}}. We have used the statistical detection methods implemented in the recently published IMGTB framework\footnote{\url{https://github.com/michalspiegel/IMGTB}}, which will be extended to also support all the fine-tunning options that we have used in this work.

Our key observations and contributions include:
\begin{itemize}
\item We have \textbf{proposed a unique way of combining the statistical and fine-tuned detection methods} using a two-way majority voting and a per-language threshold calibration.
\item We have \textbf{proposed and compared three ensemble system alternatives} to cope with multilingual machine-generated text detection (additional two in the post-deadline study).
\item We have \textbf{experienced a remarkably good performance of fine-tuned LLMs} of 7B parameters in this task. We have not used such ``a big hammer'' for the classification task before.
\item We have proposed \textbf{the best-performing single-model system} called rMistral (Mistral-7B fine-tuned in a robust way -- using both the train and dev sets and obfuscating 20\% of the train data), achieving \textbf{0.97 AUC ROC} on the test data. Although our per-language threshold calibration method would not bring the best accuracy on the test set (0.93), the threshold fixed to 1.0 (only predictions with a probability of 1, i.e. 100\% confident, are marked as machine-generated) would won the competition (accuracy of 0.97). Nevertheless, we have noticed such a threshold performance only \underline{after the deadline} and we considered the model being over-fitted (we would not submitted the results) which turned-out to be false.
\end{itemize}

\section{Background and Methodology}

For the machine-generated text detection task, three main groups of methods are nowadays used \citep{uchendu2023attribution}. The first one is a \textit{stylometric detection}, which uses linguistic features (e.g., n-grams) to differentiate between human and machine writing styles \citep{frohling2021feature, kumarage2023stylometric}. The second group is a \textit{statistical detection}, which uses statistical distribution based on a pre-trained language model (e.g., GPT2) to calculate various metrics (e.g., entropy) that can be used even without training (i.e., better generalization) to differentiate machine and human written texts \citep{mitchell2023detectgpt, hans2024spotting}. The last group is a \textit{fine-tuned detection}, which further trains an already pre-trained language model for the detection task \citep{uchendu-etal-2020-authorship, macko-etal-2023-multitude}.

In the SemEval2024 Task~8 \citep{semeval2024task8}, we have focused on the multilingual track of Subtask~A, which aimed at a binary classification to differentiate between human-written and machine-generated texts. The provided dataset (not allowing additional training data) contained the predefined splits of train, dev, and test sets. The train and dev sets officially contained 8 languages (3 languages in the dev set only), while unknown number of languages is contained in the test set.

Due to a multilingual nature of the data and our previous experience in multilingual machine-generated text detection \citep{macko-etal-2023-multitude}, we wanted to try-out something new in this shared task. Our initial idea was to experiment with a per-language ``mixture-of-experts'', which would consist of multiple models, fine-tuned in a monolingual way per each official language in the train and dev sets. Since it was expected that surprise languages will be present in the test set, we would have used an additional multilingually fine-tuned model for other languages. However, we have started the experiments only few weeks before the deadline, which gave us little time to cope with the problems such as over-fitting and hyper-parameter optimisation (shown as severe towards the deadline).

Therefore, while training these per-language models, we also started to fine-tune the Falcon-7B model \citep{falcon40b} for the machine-generated text detection task, inspired by the winning system \citep{gagiano2023prompt} of the recent ALTA 2023 shared task (although English monolingual). Since Falcon-7B is pre-trained on two languages only (English and French), we did not want to use it in a standalone way due to uncertain cross-lingual capability. Therefore, we have similarly fine-tuned the Mistral-7B model \citep{jiang2023mistral}, which is similarly sized generative model outperforming even some 13B parameters models in common benchmarks. We have not previously experimented with such a ``big hammer'' for the task; therefore, it was a new interesting experience for us. We have further combined these LLMs with statistical detectors to ensure better generalization of the system, which is described in the following sections.

\section{System Overview}

\begin{table*}
\centering
\resizebox{\linewidth}{!}{
\begin{tabular}{p{0.15\linewidth}@{}p{\linewidth}}
\hline
\textbf{System} & \textbf{Description}\\
\hline
\multirow[l]{5}{2.5cm}{$\ast$LLM2S3} & The system described in this paper. It is an ensemble using two-step majority voting for predictions, consisting of 2 LLMs (Falcon-7B and Mistral-7B) fine-tuned using the train set only, 3 zero-shot statistical methods (Entropy, Rank, Binoculars) using Falcon-7B and Falcon-7B-Instruct for calculation of the metrics, utilizing language identification and per-language threshold calibration.\\
\hline
\multirow[l]{5}{2.5cm}{PLMoE} & Our initial idea representing a per-language mixture of experts. It uses Electra-Large-Discriminator for English and XML-RoBERTa-Large for each of other languages officially present in the train and dev sets. Models for languages present in the dev set only are trained using the dev set. For unknown languages the Mistral-7B fine-tuned using the whole train set is used.\\
\hline
\multirow[l]{4}{2.5cm}{rLLM2S3} & The same ensemble system as LLM2S3; however, the LLMs are fine-tuned using both the train and dev sets. Also, to increase the robustness of the system, we have obfuscated 20\% of the train samples during fine-tuning, by using HomoglyphAttack and inserting zero-width-joiner character, inspired by our recent work \citep{macko2024authorship}.\\
\hline\hline
\multirow[l]{4}{2.5cm}{rLLM2B-ES} & The post-deadline ensemble system similar to rLLM2S3; however, the Llama-2-7B is used instead of Falcon-7B and Binoculars is used solely in the statistical part (instead of a combination of 3 methods). Moreover, the fine-tuning process used the early stopping mechanism to alleviate the over-fitting.\\
\hline
\multirow[l]{5}{2.5cm}{LLM2B1} & The post-deadline ensemble system using the original LLM2S3 fine-tuned Falcon-7B and Mistral-7B models; however, classification thresholds are not calibrated, but only predictions with a probability of 1 (i.e., 100\% confident) are marked as machine-generated. Such predictions are combined with Binoculars zero-shot prediction using the per-language threshold calibration.\\
\hline
\end{tabular}
}
\caption{Description of system alternatives. The main system described in this paper is denoted by $\ast$. The last two alternatives were evaluated post-deadline.}
\label{tab:alternatives}
\vspace{-3mm}
\end{table*}

Our best system (see \figurename~\ref{fig:architecture}) combines the predictions of two fine-tuned LLMs (Falcon-7B and Mistral-7B) with the selected statistical metrics (Entropy, Rank, Binoculars) by using \textbf{a two-step majority voting}. Firstly, a single majority-voted prediction results out of the three statistical metrics. Then, the final majority-voted prediction is a combination of the previous one with the Falcon and Mistral predictions.

Each prediction uses a separate classification-decision threshold, which is applied on prediction probabilities and statistical metrics. These \textbf{thresholds are calibrated in a per-language way}, meaning that separate thresholds are used for each language officially present in the train and dev sets, plus an additional threshold for unknown languages (i.e., not officially present in the train and dev sets). The thresholds are calibrated based on the machine-class prediction probabilities and statistical metrics for samples in the train and dev sets combined. The calibration maximized the difference between true positive rate (TPR) and false positive rate (FPR) based on the ROC (receiver operating characteristic) curve. The texts with probabilities (or statistical metrics) outreaching the thresholds are considered machine-generated, otherwise they are considered human-written. The thresholds are saved and used for prediction of test samples.

Due to unknown languages in the test set and using the per-language threshold calibration, we have utilized the FastText\footnote{\url{https://pypi.org/project/fasttext-langdetect/}} \textbf{language identification}. Since it is not fully accurate, we have used such language information only if the prediction probability was greater than 0.5, otherwise the language was handled as unknown.

As mentioned, the system includes \textbf{two fine-tuned LLMs}, namely Falcon-7B and Mistral-7B.
For the fine-tuning process, we have used a parameter efficient fine-tuning (\textbf{PEFT}) technique called \textbf{QLoRA} \citep{dettmers2023qlora} to minimize the computational costs of our system training.

To enhance the system performance generalization, we have integrated a \textbf{statistical part} of the system, which is based on the three statistical metrics, namely Entropy \citep{10.5555/3053718.3053722}, Rank \citep{gehrmann2019gltr}, and recently proposed Binoculars \citep{hans2024spotting}. The statistical metrics are calculated using the Falcon-7B as a base model. Since Binoculars requires two models, it uses also Falcon-7B-Instruct (as a performer model).

Besides the described best submitted system, we have tried multiple system alternatives, which are briefly described in Table~\ref{tab:alternatives}. In addition to those ensembles, we have evaluated single detectors, namely Falcon, Mistral, S5 (a combination of 5 statistical metrics -- likelihood, entropy, rank, log-rank, and llm-deviation), and Binoculars. After the deadline, we have also finished fine-tuning of Llama-2-7B and retrained the detectors using the early stopping (patience of 5) to prevent over-fitting. Also, when knowing the gold labels of the test set, we have evaluated various combinations of the trained detectors to see whether we have done the right decision for the submission.

\section{Experimental Setup}

For the experimental purpose, we have used the defaults splits of the provided dataset, namely the train and dev sets in the pre-deadline experiments, and the gold labels of the test set for the post-deadline evaluation of the pre-deadline system alternatives. The main system described in this paper uses only the train set in the training process; however, uses both the train and dev sets for the classification threshold calibration. Some of the system alternatives used both the train and dev sets in the training process, as described in Table~\ref{tab:alternatives}.

As the key evaluation metric in the shared task is \textbf{accuracy}, we have also used this metric for the preliminary system evaluation and selection of the alternative for submission. Since classification task is sensitive to the used classification threshold, we have also used \textbf{AUC ROC} (area under curve of the receiver operating characteristic) as a threshold independent metric, providing better information about the classification capability.

For the fine-tuning process, we have used the official baseline script\footnote{\url{https://github.com/mbzuai-nlp/SemEval2024-task8/blob/main/subtaskA/baseline/transformer_baseline.py}}, modified to export machine-class prediction probabilities in addition to the predictions. Since, it was not clear which version of the XLM-RoBERTa model was marked as a baseline in the multilingual track (with the known accuracy of 0.72), we have trained both the base (\textit{XLM-R-B}) and large (\textit{XLM-R-L}) versions. In addition, we have also included mDeBERTa-v3-base (\textit{mDeBERTa}) model in our baselines, since it performed the best in our previous work \citep{macko-etal-2023-multitude}.

To perform per-language models fine-tuning, we have used the source field of the train and dev data to select data only for the specific language. Other parameters of the fine-tuning process remained unchanged. The FastText language identification is used for a prediction, which uses the machine-class probability of the corresponding language-specific model.

The used QLoRA PEFT fine-tuning process used the binary cross entropy with logits for loss calculations and 4-bit quantization using BitsAndBytes\footnote{\url{https://pypi.org/project/bitsandbytes}}. The LoRA configuration\footnote{\url{https://pypi.org/project/peft}} used an \textit{alpha} of 16, a \textit{dropout} of 0.1, \textit{r} of 64, and the \textit{task type} of sequence classification. Unlike the baseline fine-tuning, this version used half-precision training, gradient accumulation of 4 steps, and evaluation each 1,000 steps. Other parameters were the same.

Due to time constraints, we have not done any hyper-parameter optimization; thus, further improvements of the system are very likely possible.

\section{Results}

The experimental results are provided in Table~\ref{tab:results}. It must be noted that the results in the bottom part of the table are not part of the competition, since those experiments were performed after the submission deadline of the shared task. Also, the performance results using the test set were not known before the deadline; gold labels has been released only afterwards. Therefore, the design decisions could be made purely using the dev set.

\begin{table}[!t]
\centering
\resizebox{\linewidth}{!}{
\begin{tabular}{ll|cc|cc}
\hline
& & \multicolumn{2}{c|}{\textbf{Accuracy}} & \multicolumn{2}{c}{\textbf{AUC ROC}}\\
& \textbf{System} & \textbf{Dev} & \textbf{Test} & \textbf{Dev} & \textbf{Test}\\
\hline
\multirow[l]{3}{*}{\textbf{Baselines}}
& XLM-R-B & 0.7158 & 0.7935 & 0.8262 & 0.9040\\
& XLM-R-L & 0.7275 & 0.8841 & 0.8187 & 0.9063\\
& mDeBERTa & 0.6968 & 0.8943 & 0.7952 & 0.9832\\
\hline
\multirow[l]{3}{2cm}{\textbf{System\\ Alternatives}}
& $\ast$LLM2S3$\bullet$ & 0.9035 & 0.9501 & N/A & N/A\\
& PLMoE$\bullet$ & \textcolor{lightgray}{0.9878} & 0.5819 & \textcolor{lightgray}{0.9943} & 0.6268\\
& rLLM2S3$\bullet$ & \textcolor{lightgray}{0.9965} & 0.9560 & N/A & N/A\\
\hline
\multirow[l]{8}{2cm}{\textbf{Ablation}\\ \textbf{Study}}
& Falcon & 0.8043 & 0.9102 & 0.8775 & 0.9492\\
& Mistral & 0.8560 & 0.9027 & 0.9138 & 0.9579\\
& rFalcon & \textcolor{lightgray}{0.9905} & 0.8843 & \textcolor{lightgray}{0.9991} & 0.9395\\
& rMistral & \textcolor{lightgray}{\textbf{0.9980}} & 0.9268 & \textcolor{lightgray}{\textbf{0.9997}} & 0.9713\\
& S3$\bullet$ & 0.7248 & 0.8328 & N/A & N/A\\
& S5$\bullet$ & 0.5880 & 0.4763 & N/A & N/A\\
& Binoculars & 0.5430 & 0.7979 & 0.6304 & 0.8777\\
& Binoculars$\bullet$ & 0.6240 & 0.8434 & 0.6304 & 0.8777\\
\hline\hline
\multirow[l]{7}{2cm}{\textbf{Post-Deadline Study}}
& PLMoE-ES$\bullet$ & \textcolor{lightgray}{0.9885} & 0.8417 & \textcolor{lightgray}{0.9947} & 0.9635\\
& Llama-2 & 0.7335 & 0.7587 & 0.9342 & 0.7400\\
& rLlama-2 & \textcolor{lightgray}{0.8903} & 0.8907 & \textcolor{lightgray}{0.8416} & 0.9400\\
& rLlama-2-ES & \textcolor{lightgray}{0.9838} & 0.8805 & \textcolor{lightgray}{0.9960} & 0.9108\\
& rFalcon-ES & \textcolor{lightgray}{0.9410} & 0.8672 & \textcolor{lightgray}{0.9872} & 0.9503\\
& rMistral-ES & \textcolor{lightgray}{0.9863} & 0.9412 & \textcolor{lightgray}{0.9984} & \textbf{0.9834}\\
& rLLM2B-ES$\bullet$ & \textcolor{lightgray}{0.9915} & 0.9700 & N/A & N/A\\
& LLM2B1 & 0.8668 & \textbf{0.9708} & N/A & N/A\\
& rMistral1 & \textcolor{lightgray}{0.9975} & 0.9675 & \textcolor{lightgray}{\textbf{0.9997}} & 0.9713\\
\hline
\end{tabular}
}
\caption{Detection performance evaluated using the dev (pre-deadline) and test (post-deadline) splits separately. The main system described in this paper is denoted by $\ast$, the systems using the per-language threshold calibration are denoted by $\bullet$, systems using fixed threshold of 1.0 are denoted by ``$1$''. ``-ES'' denotes using of early-stopping mechanism to prevent over-fitting. ``N/A'' denotes not available values due to prediction-based majority voting (i.e., no probabilities to calculate AUC ROC). The gray color denotes unrepresentative performance values due to training on the dev set.}
\label{tab:results}
\vspace{-5mm}
\end{table}

Due to high accuracy and high AUC ROC metrics using the dev set, we considered \textit{rFalcon} and \textit{rMistral} over-fitted; therefore, we decided not to submit our \textit{rLLM2S3} system. This turned-out to be a mistake, since it performed slightly better than the submitted \textit{LLM2S3} on the test set. On the other hand, our suspicion of over-fitting \textit{PLMoE} (due to the similar observations) turned-out to be valid, since it performed much worse using the test set. Therefore, it seems that per-language monolingually fine-tuned (i.e., lower amount of samples) models require optimization of hyper-parameters to prevent over-fitting and to better generalize to unseen texts.

As an ablation study, we also provide the results for individual components of our system alternatives. As the results show, the ensembling into more complex systems of \textit{LLM2S3} and \textit{rLLM2S3} helped generalization of the classification performance. Individual methods would not outperform the submitted system.

\subsection{Post-Deadline Study}

In the post-deadline experiments (already knowing the gold labels of the test set for evaluation), we have finished Llama-2-7B model fine-tuning and retraining all three robust-version LLMs using the early stopping (to minimize the over-fitting). The results revealed that the \textit{rLlama-2} model does not suffer by over-fitting as much. Based on the test set evaluation and by examining various combinations, the retrained \textit{rLlama-2-ES} and \textit{rMistral-ES} seemed like good candidates to combine with Binoculars (\textit{rLLM2B-ES}), outperforming the winning system in the competition.

Early stopping helped a lot in boosting performance generalization (i.e., reducing over-fitting) of our per-language mixture-of-experts ensemble system (\textit{PLMoE-ES}), achieving one of the highest AUC ROC using the test set. Nevertheless, in the accuracy as an official metric, it would not outperform the other system alternatives.

In addition, we have noticed that optimal thresholds for fine-tuned LLMs are often set to 1.0 by using purely the dev set samples machine-class probabilities. Therefore, we have fixed the thresholds to 1.0 for the \textit{LLM2B1} system (containing only models we have trained before the deadline), meaning that the machine-class predictions of the LLMs are used only when having 100\% confidence (otherwise considered human-written). Such predictions, when combined with Binoculars, achieved even higher performance using the test set data (0.9708). Thus, we had such a system trained before the deadline; however, we have not noticed such a threshold bringing the best performance in time. Moreover, when looking at the accuracy for the dev set, we do not see why we would select such a system for the submission. It can be just a coincidence that it performs so well using the test set data. Further experiments are required to examine this phenomenon using independent out-of-distribution data.

Also, even when our \textit{rLLM2B-ES} system alternative or the \textit{rMistral1} single-model system would won the competition, we are now not sure that we would be confident enough (about not being over-fitted) to submit it as the final system without evaluation on the external dataset. Thus, we have submitted best what we could at the time.

\subsection{Per-language Analysis}

\begin{figure}[!t]
    \centering
    \includegraphics[width=\linewidth, trim=0.2cm 0 1.5cm 0, clip]{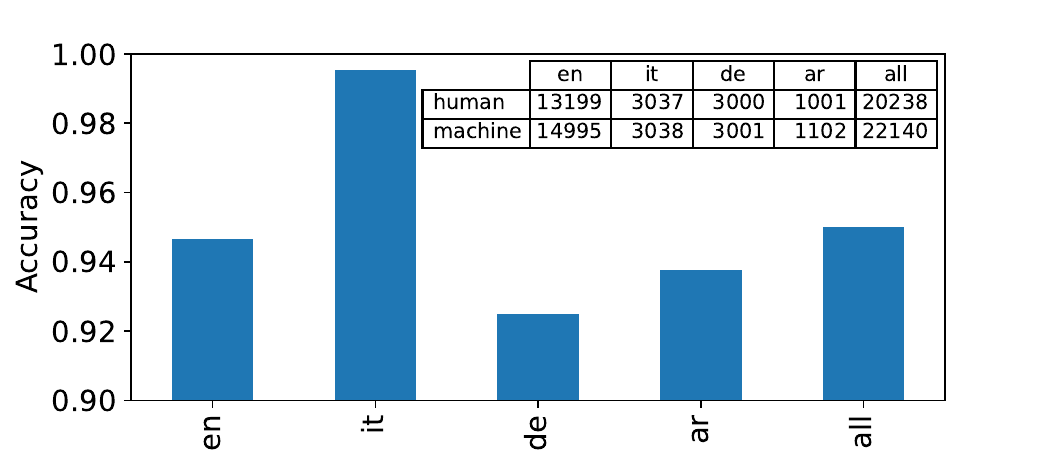}
    \caption{Per-language test-set performance (\textit{it} is a surprise language, \textit{de} and \textit{ar} are in the dev set only). Axis scale for Accuracy is shown from 0.9 to 1.0. The per-class samples counts are provided in the top-right table.}
    \label{fig:languages}
\end{figure}

For a deeper insight of the proposed system (\textit{LLM2S3}) performance, we have performed an analysis per each language identified in the test set. The results are provided in \figurename~\ref{fig:languages}. Interesting is that it achieved the highest accuracy for the Italian surprise language (\textit{it}). Lower accuracy is evident for German and Arabic languages, although present in the dev set. It must be noted that this version of the system was not trained using the dev set, only the classification threshold calibration used such data. Therefore, the robust versions of system alternatives are expected to provide higher performance especially in those languages.

\section{Conclusion}

To cope with the problem of multilingual, multidomain, and multigenerator machine-generated text detection, we have proposed an ensemble system using 2 LLMs (Falcon-7B and Mistral-7B) fine-tuned for the binary sequence classification task. We have further combined the predictions with the statistical metrics of Entropy, Rank, and Binoculars using a two-stage majority voting. The classification thresholds in our system have been calibrated in a per-language manner, for which we have utilized the FastText language identification.
A combination of fine-tuned LLMs and statistical detection seems to be the right way to cope with generalization of the detection performance. Out of the evaluated single-model systems, Mistral-7B is the best candidate for fine-tuning, which by itself can bring a remarkable classification performance. Further improvements of the system could be easily achievable by hyper-parameters optimization, which we have not done in the submitted system due to lack of time.

\section*{Acknowledgements}

This work was partially supported by the projects funded by the European Union under the Horizon Europe: \textit{AI-CODE}, a project funded by the European Union under the Horizon Europe, GA No. \href{https://cordis.europa.eu/project/id/101135437}{101135437}, and \textit{VIGILANT}, GA No. \href{https://doi.org/10.3030/101073921}{101073921}.
Part of the research results was obtained using the computational resources procured in the national project \textit{National competence centre for high performance computing} (project code: 311070AKF2) funded by European Regional Development Fund, EU Structural Funds Informatization of Society, Operational Program Integrated Infrastructure.

\bibliography{anthology, custom}

\appendix

\section{Computational Resources}
\label{sec:A}

For experiments regarding model fine-tuning and inference processes, we have used $1\times$ NVIDIA GeForce RTX 3090 24GB GPU and $1\times$ A100 40GB GPU, cumulatively consuming around 10,000 GPU-core hours. For combining the results and analysis, we have used Jupyter Lab running on 4 CPU cores, without the GPU acceleration.

\end{document}